\documentclass{vgtc}                          

\usepackage{times}                     %
\usepackage{tabu}                      %
\usepackage{booktabs}                  %
\usepackage{lipsum}                    %
\usepackage{mwe}                       %
\usepackage{url}
\usepackage{xcolor}
\usepackage{xspace}
\usepackage{enumitem}
\usepackage{wrapfig}

\usepackage{mathptmx}                  %

\usepackage{balance}

\onlineid{0}

\vgtccategory{Research}

\vgtcinsertpkg

\title{Dimensionality Reduction Meets Network Science: \\Sensemaking on UMAP's kNN Graph  }
\author{Duen Horng (Polo) Chau\thanks{e-mail: polochau@apple.com}\\ %
        \scriptsize Apple %
\and Donghao Ren\thanks{e-mail: donghao@apple.com}\\ %
     \scriptsize Apple %
\and Fred Hohman\thanks{e-mail: fredhohman@apple.com}\\ %
     \scriptsize Apple %
\and Dominik Moritz\thanks{e-mail: domoritz@apple.com}\\ %
     \scriptsize Apple %
}

\teaser{
\centering
\includegraphics[width=\linewidth]{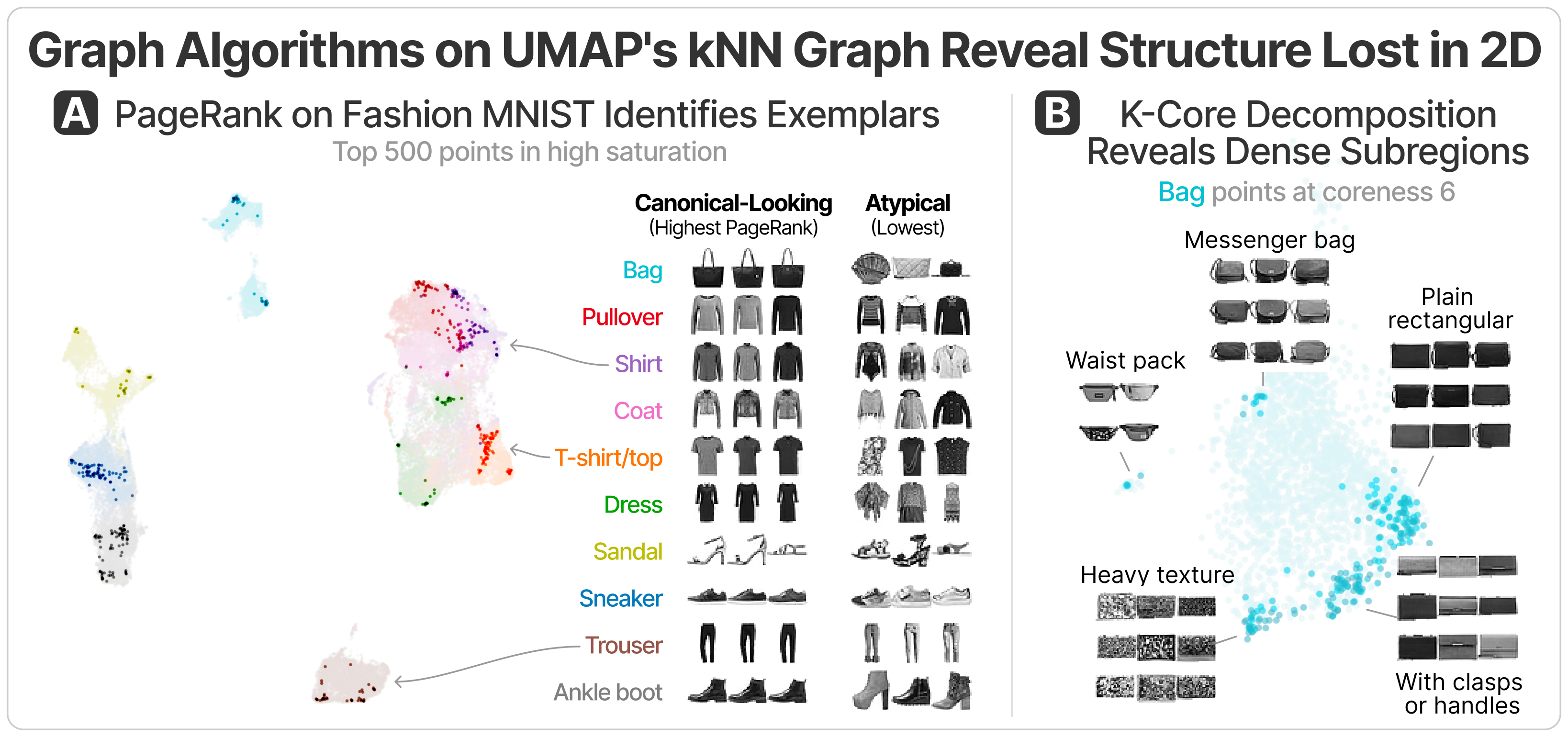}
\vspace{-20pt}
\caption{Standard graph algorithms applied to UMAP's internal $k$NN graph reveal structure lost in 2D scatter plot layouts. 
\textbf{A.}~PageRank on the Fashion MNIST $k$NN graph identifies representative data points. The highest-scoring points exhibit prototypical appearances, while the lowest-scoring points display atypical variations. Top 500 points shown in high saturation. 
\textbf{B.}~k-core decomposition reveals distinct sub-categories in the ``bag'' class (e.g., messenger bags, waist packs, heavy textures) through filtering for a coreness of 6 from the dense mass of points in the 2D scatter plot layout. Grayscale images inverted for clarity.
}
\label{fig:crownjewel}
}

\abstract{
While UMAP is widely used for exploring high-dimensional data, typical workflows focus on its lower-dimensional embedding, largely overlooking the rich $k$-nearest-neighbor (kNN) graph that UMAP constructs internally. 
This graph encodes the data manifold in its original high-dimensional space, \textit{before} the distortion that UMAP's 2D projection introduces.
We demonstrate the untapped potential of this internal representation, showing how standard graph algorithms applied to this graph enhance data sensemaking:
(1)~\textit{PageRank} identifies representative data points,
(2)~\textit{k-core decomposition} reveals dense core regions versus sparse periphery, and
(3)~\textit{clustering coefficient} detects tight-knit neighborhoods with highly-similar data points.
Through quantitative and qualitative evaluation on MNIST and Fashion MNIST, we show that these graph-based analyses are not only practical but also competitive with or complementary to purpose-built methods (e.g., $k$-medoids for exemplar selection, HDBSCAN for density-based clustering).
}


\begin{document}

\firstsection{Introduction}

\maketitle

UMAP~\cite{2018arXivUMAP} is among the most widely used tools for visually exploring high-dimensional data.
Yet the 2D scatter plot output by UMAP is typically treated as the sole analytical artifact~\cite{espadoto2019toward,jeon2025stop}.

\begin{figure}[t]
\centering
\includegraphics[width=\linewidth]{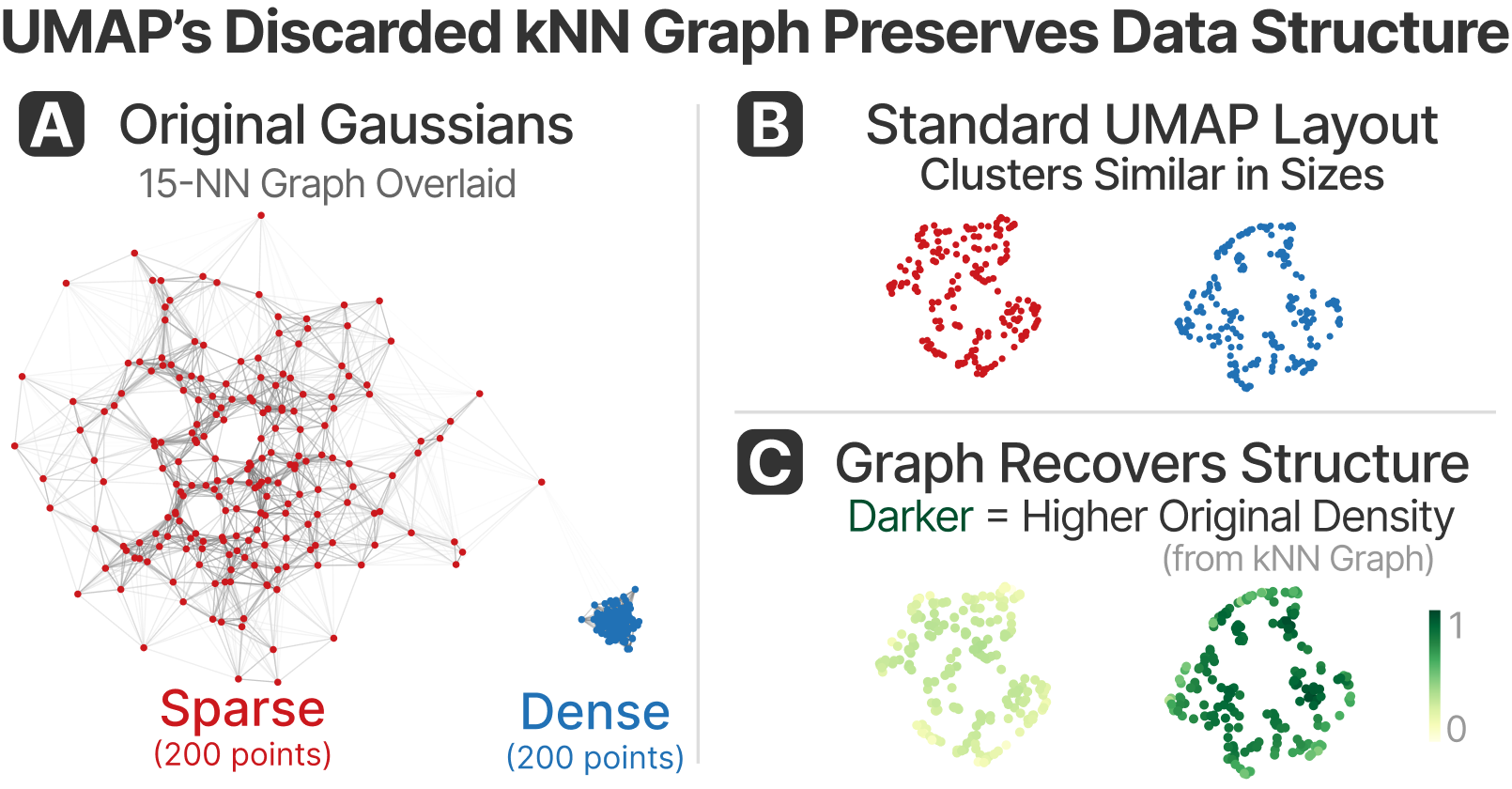}
\caption{\textbf{UMAP's 2D layout discards structural information preserved in its intermediate kNN graph.}
\textbf{A.}~Original data with two Gaussian clusters of same point counts: one \textcolor[HTML]{BA2F29}{sparse} and one \textcolor[HTML]{3A6FB0}{dense}, with 15-NN graph overlaid (shorter edges darker).
\textbf{B.}~Standard 2D UMAP normalizes local distances; clusters appear similar sized.
\textbf{C.}~Same layout, colored by graph-derived local density (inverse mean kNN distance in original space; darker\,=\,denser),
recovering structure lost in projection. 
The graph also encodes centrality, connectivity hierarchy, and local cohesion (\cref{sec:approach}).}
\vspace{-10pt}
\label{fig:what-graph}
\end{figure}

\textbf{UMAP's discarded kNN graph.}
Before producing a 2D layout, UMAP builds a weighted directed graph that models the data manifolds' local geometry (\cref{fig:what-graph}). 
For each point, it finds $k$ nearest neighbors in high-dimensional space, then applies a density-adaptive normalization:
each point's bandwidth~$\sigma_i$ is calibrated to its local density, transforming raw distances into \textit{membership strengths} in $[0,1]$ that are comparable across sparse and dense regions~\cite{2018arXivUMAP}.
Every point has exactly $k$ outgoing edges, but in-degree varies: points deemed similar by many others receive more incoming edges, while dissimilar ones are nominated by few.
We call this the \textbf{kNN graph} for short.
This $k$NN graph encodes the manifold's connectivity far more faithfully than the 2D scatter plot layout that UMAP subsequently optimizes from it~\cite{coenenunderstanding,jeon2025stop}---yet after layout optimization, the graph is typically set aside.
We argue it should be retained as a \textbf{first-class} analytical resource.

\textbf{Sensemaking from kNN graph.}
Because the $k$NN graph faithfully reflects the high-dimensional manifold, standard graph algorithms can be applied to answer sensemaking questions---about representativeness, density structure, and local cohesion---that the 2D scatter plot alone cannot.

\smallskip
\noindent
Our work makes two major contributions:
\begin{enumerate}[nosep, leftmargin=*]
\item \textbf{Elevating UMAP's kNN graph to a first-class analytical resource,}
bridging dimensionality reduction and network science. 
Instead of treating this graph as a disposable intermediate,
this perspective enables direct application of standard graph algorithms 
to leverage the high-dimensional manifold encoded \textit{before} projection distortion, enabling complementary sensemaking beyond what 2D scatter plot layouts provide: 
(1)~\textit{PageRank}~\cite{brin1998anatomy} identifies globally representative points via transitive centrality: a node ranks high when many high-ranking nodes nominate it as a neighbor;
(2)~\textit{k-core decomposition}~\cite{batagelj2003an} reveals dense core regions versus sparse periphery; 
and
(3)~\textit{clustering coefficient}~\cite{watts1998collective} detects tight-knit micro neighborhoods with highly-similar data points.

\item \textbf{Quantitative and qualitative evidence of effectiveness.}
On the MNIST and Fashion MNIST datasets, 
graph-based analyses prove competitive or complementary to purpose-built methods (e.g., $k$-medoids~\cite{park2009simple} for exemplar selection, HDBSCAN for density-based clustering):
PageRank-selected data points achieve superior class balance to $k$-medoids and rival on representativeness and downstream classification accuracy;
k-core decomposition reveals a graduated subgraph hierarchy that discrete cluster labels (e.g., from HDBSCAN~\cite{mcinnes2017hdbscan}) cannot capture;
and clustering coefficient reveals distinct micro-clusters where a point's nearest neighbors also nominate one another.

\end{enumerate}

\section{Related Work}
\textbf{Interpreting dimensionality reduction (DR).}
Surveys of DR techniques have extensively cataloged quality metrics, distortion measures, and layout enrichment strategies~\cite{nonato2019multidimensional,espadoto2019toward}, and recent work has highlighted persistent misinterpretation of DR outputs~\cite{jeon2025stop,coenenunderstanding}.
These efforts focus on improving or qualifying the 2D layout itself.
We take a complementary path: rather than refining layout interpretation, we operate directly on the pre-projection kNN graph that UMAP~\cite{2018arXivUMAP} constructs as an intermediate representation.

\textbf{Clustering and selection on embeddings.}
Data analysis workflows commonly project data with UMAP and then apply clustering or selection on the low-dimensional output with HDBSCAN~\cite{mcinnes2017hdbscan,becht2019dimensionality,allaoui2020considerably} or $k$-medoids~\cite{park2009simple}.
These methods analyze the \emph{projected} output.
Our approach instead operates on UMAP's pre-projection kNN graph, which encodes the high-dimensional manifold before layout optimization distorts it.

\textbf{kNN graphs in manifold learning.}
Manifold learning methods such as Laplacian Eigenmaps~\cite{belkin2003laplacian} and UMAP~\cite{2018arXivUMAP} build explicit kNN graphs as an intermediate step, yet discard them after computing the embedding.
We argue this graph should be retained, and show that classical graph algorithms---PageRank~\cite{brin1998anatomy}, k-core decomposition~\cite{batagelj2003an}, clustering coefficient~\cite{watts1998collective}---applied to it yield sensemaking capabilities complementary to the 2D scatter plot.
Our work also differs from domain-specific workflows (e.g., scanpy~\cite{wolf2018scanpy}) that build a separate graph for community detection~\cite{traag2019louvain}; we instead reuse UMAP's own graph and apply complementary per-point scoring algorithms.

\section{Graph Algorithms on UMAP's kNN Graph}
\label{sec:approach}

The UMAP $k$NN graph exhibits distinctive structural properties---fixed out-degree~$k$, variable in-degree, and density-adaptive edge weights.
These properties suggest graph algorithms for three sensemaking questions:
\textit{PageRank} to identify representative data points (\cref{sec:pagerank}), \textit{k-core} decomposition to reveal dense regions (\cref{sec:kcore}), and the \textit{clustering coefficient} to detect cohesive local neighborhoods (\cref{sec:cc}). 
We evaluate our approach on MNIST~\cite{lecun1998gradient} and Fashion MNIST~\cite{xiao2017fashion} (each containing 60{,}000 images across 10 classes), both widely established benchmarks for dimensionality reduction~\cite{espadoto2019toward,2018arXivUMAP}. 
We compare our methods against $k$-medoids~\cite{park2009simple} and HDBSCAN~\cite{mcinnes2017hdbscan}, standard baselines in this domain~\cite{charikar2023simple,kwon2025benchmarking,allaoui2020considerably}.

\subsection{PageRank for Representative Selection}
\label{sec:pagerank}

When exploring a UMAP projection, a natural first question is \textit{``which points best represent this data?''}
Selecting representative data points, or \textit{exemplars}, helps build a mental model of the data, yet existing approaches are often ill-suited to this task.
A common approach is to choose the point nearest to a cluster's low-dimensional centroid (e.g., in 2D), relying on UMAP's distorted geometry; the centroid of a stretched or compressed cluster may fall in an unrepresentative region.

The kNN graph offers a natural alternative. Its in-degree encodes a point's centrality in the manifold: a point with high in-degree is frequently nominated as a nearest neighbor by others, meaning many data points consider it representative of their local region. PageRank~\cite{brin1998anatomy}---originally developed to identify authoritative web pages based on hyperlink structures---amplifies this signal transitively: a node scores high not only because many nodes point to it, but because \textit{those nodes} are themselves highly pointed-to.\footnote{PageRank correlates strongly with weighted in-degree ($\rho=0.93$) but captures transitive structure beyond it: ranking points by the gap between their PageRank and in-degree ranks, the top $1\%$ (points most boosted by PageRank) have $4.3\times$ higher mean neighbor-PageRank than the bottom $1\%$.
We adopt PageRank as a principled, parameter-free formulation with decades of tooling support, and note that weighted in-degree may serve as an alternative for latency-sensitive work.}

\begin{figure}[t]
\centering
\includegraphics[width=0.9\linewidth]{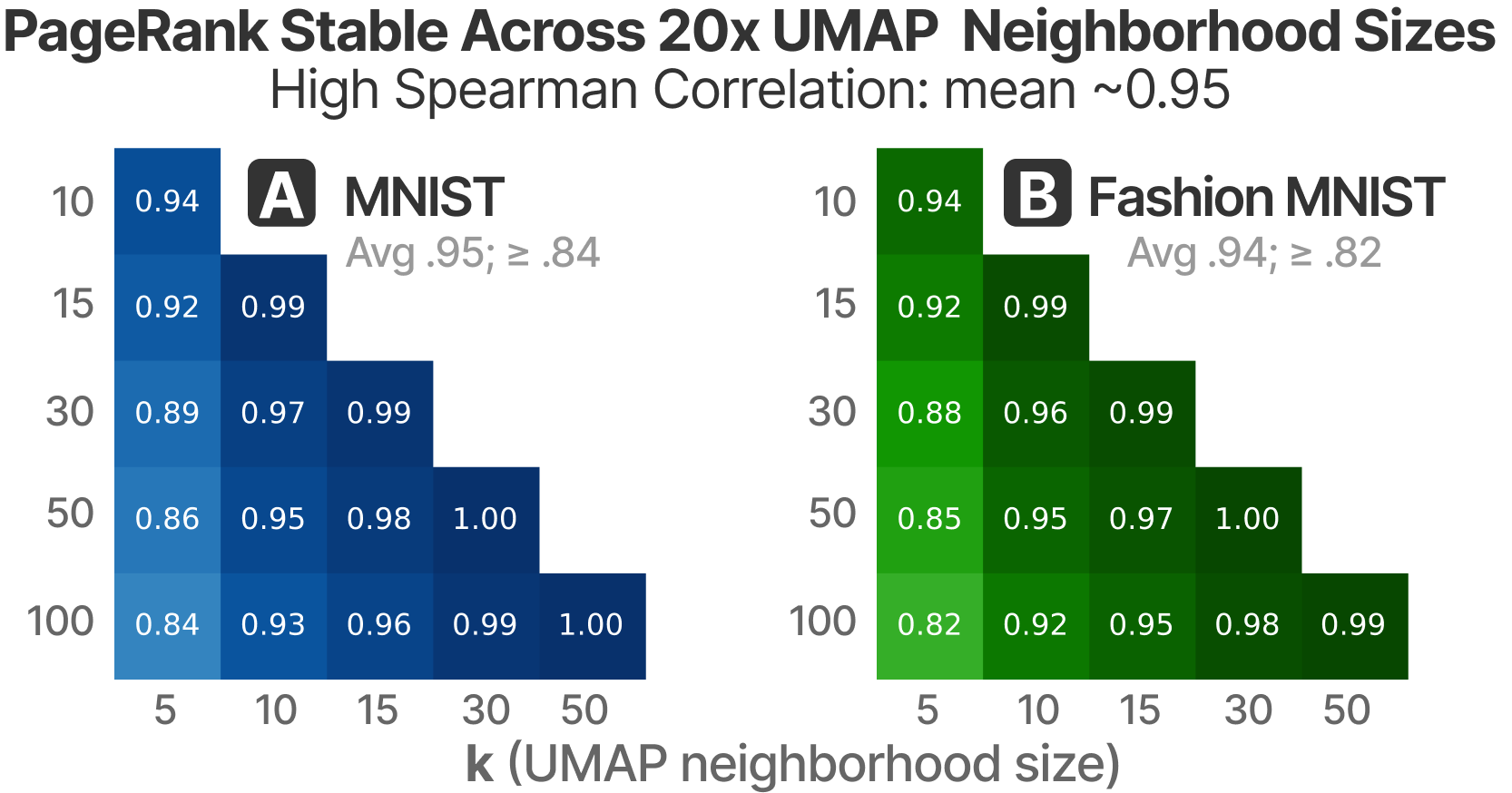}
\caption{Data points' PageRank rankings achieve high Spearman rank correlations across both datasets (mean $\sim0.95$) over $20{\times}$ range of UMAP neighborhood sizes (5--100), confirming PageRank-selected representatives are not neighborhood-scale artifacts.}
\label{fig:pagerank-stability}
\vspace{-10pt}
\end{figure}

\subsubsection{Evaluation: PageRank}

\textbf{Stability of representatives across neighborhood sizes.}
We run PageRank on the weighted kNN graph using UMAP's density-adaptive membership strengths as edge weights (default damping factor: $0.85$).
Because each point's  
bandwidth~$\sigma_i$ is calibrated so that its total 
outgoing weight is approximately constant regardless of local density~\cite{2018arXivUMAP},
PageRank reflects topological centrality rather than merely recapitulating local density.
Since $k$ (UMAP's \texttt{n\_neighbors}) controls graph construction, governing each point's out-neighbors, we tested whether PageRank rankings are sensitive to this choice.
We varied $k$ across $\{5, 10, 15, 30, 50, 100\}$ (a $20{\times}$ range) and recomputed PageRank for each graph.
As shown in \Cref{fig:pagerank-stability}, on both datasets, pairwise Spearman rank correlations remained high (MNIST: $\rho \geq 0.84$, mean~$0.95$; Fashion MNIST: $\rho \geq 0.82$, mean~$0.94$); moreover, $50$--$60\%$ of each top-$s$ selection ($s\le1000$) recurred across all six $k$ values (mean pairwise Jaccard $0.62$--$0.71$), confirming that the selected representatives are not artifacts of a particular neighborhood scale. We use $k=15$, which is also UMAP's default, for subsequent experiments.
\Cref{fig:crownjewel}A shows that the highest-scoring points in each Fashion MNIST class exhibit prototypical appearances, while the lowest-scoring points display atypical variations. 

\begin{figure}[t]
\centering
\includegraphics[width=0.9\linewidth]{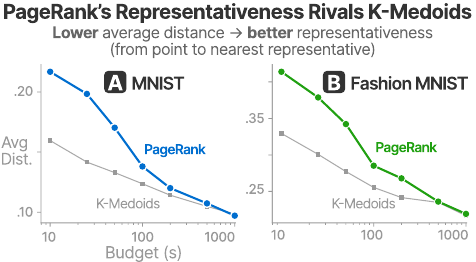}
\vspace{-7pt}
\caption{PageRank rivals $k$-medoids on representativeness despite not optimizing the distance objective. Representativeness measured as mean cosine distance from each data point to its nearest selected representative (lower is better).}
\label{fig:pagerank-representativeness}
\vspace{-13pt}
\end{figure}

\textbf{Representativeness and class balance.}
We evaluated exemplar quality along two axes:
(1)~representativeness (mean cosine distance from each point to its nearest selected representative; lower is better) and
(2)~class balance (\emph{Jensen-Shannon divergence}, JSD, between the selected and global class distribution; lower is better).
For a given budget $s$, 
PageRank selects the $s$ highest-scoring points; 
$k$-medoids directly optimizes this distance objective, selecting $s$ medoids (actual data points) that minimize total within-cluster distance\footnote{\texttt{sklearn\_extra.cluster.KMedoids} (``alternate'' variant); exact PAM is intractable at $n{=}60{,}000$.}---an objective PageRank does not optimize at all. 

\begin{figure}[b]
\centering
\includegraphics[width=0.9\linewidth]{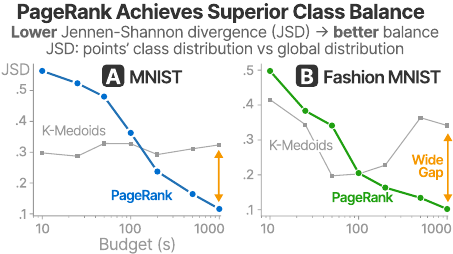}
\vspace{-10pt}
\caption{PageRank's representatives achieve superior class balance vs $k$-medoids, with a gap that rapidly widens after selecting only 200 top-scoring points ($0.3\%$ of data).}
\label{fig:pagerank-class-balance}
\end{figure}

On \textit{representativeness} alone, as shown in \Cref{fig:pagerank-representativeness}, 
$k$-medoids held a modest advantage at tiny budgets (10, 25, 50), 
but PageRank's topological selection rapidly approached $k$-medoids' as the budget $s$ reached 100, still a small budget ($0.17\%$ of data).  
Crucially, as shown in \Cref{fig:pagerank-class-balance}, this came with a \textbf{significant advantage in class balance}:
at $s \geq 200$ ($0.3\%$ of data), PageRank produced substantially lower JSD than $k$-medoids, and the gap widened rapidly with $s$.
Notably, $k$-medoids' JSD worsened as $s$ grew on Fashion MNIST (from $0.20$ at $s{=}50$ to $0.36$ at $s{=}500$): because it optimizes geometric distance, additional medoids are disproportionately allocated to high-variance, spread-out classes, pulling the selection further from the global distribution. PageRank's selection on the $k$NN graph produces significantly more proportional class representation, with JSD decreasing as $s$ grows. 

We chose k-medoids over 
HDBSCAN because it directly optimizes the distance objective and selects actual data points. 
HDBSCAN's \texttt{exemplars\_}~\cite{mcinnes2017hdbscan} are not a controllable selection mechanism: 
across a range of \texttt{min\_cluster\_size} values ($25$--$5{,}000$), HDBSCAN produced $9{,}000$--$43{,}000$ exemplars ($15$--$72\%$ of the data) with no way to specify a target budget~$s$.
By contrast, PageRank provides a continuous ranking that can be thresholded at any~$s$.

\begin{figure}[tb]
\centering
\includegraphics[width=0.90\linewidth]{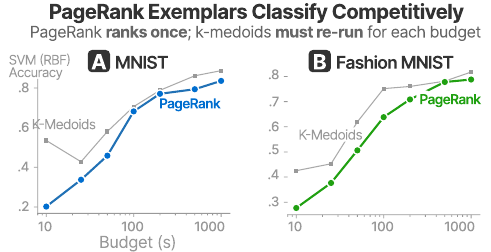}
\vspace{-10pt}
\caption{PageRank exemplars classify competitively vs $k$-medoids. SVM-RBF trained on each method's selected representatives, evaluated on all remaining points. PageRank ranks once, but $k$-medoids must re-run for each budget.}
\label{fig:pagerank-downstream-classification}
\vspace{-10pt}
\end{figure}

\textbf{Downstream classification.}
To test whether selected representatives preserve useful information, we 
trained an SVM with RBF kernel solely on the selected points and used it to classify all remaining points in the dataset,
based on evaluation methodology from the dimensionality reduction literature~\cite{wang2021understanding,espadoto2019toward}.
While $k$-medoids' spatially optimized selections gave it some edge on SVM accuracy, PageRank remained competitive---within $2$--$3$ percentage points at $s \geq 500$ on Fashion MNIST (\cref{fig:pagerank-downstream-classification}).
On MNIST, both methods reached ${\sim}84$--$89\%$ at $s{=}1000$.
This competitiveness is significant: PageRank computes a single global ranking by reusing the kNN graph UMAP has already built, which can then be thresholded at any budget~$s$ without recomputation.
By contrast, $k$-medoids must be re-run for each choice of~$s$.

\subsection{k-Core Decomposition for Density Hierarchy}
\label{sec:kcore}

Understanding which points lie in the dense core of a cluster versus its sparse periphery is critical for assessing data quality, identifying outliers, and gauging confidence in cluster membership~\cite{jeon2025stop}.
HDBSCAN~\cite{mcinnes2017hdbscan} 
assigns each point a discrete cluster label but is not designed for distinguishing centrality \emph{within} a cluster: a large cluster may contain both core and periphery members, all with the same label.

\setlength{\intextsep}{0pt}   %
\setlength{\columnsep}{10pt} %
\begin{wrapfigure}{r}{0.45\linewidth} %
  \centering
  \includegraphics[width=\linewidth]{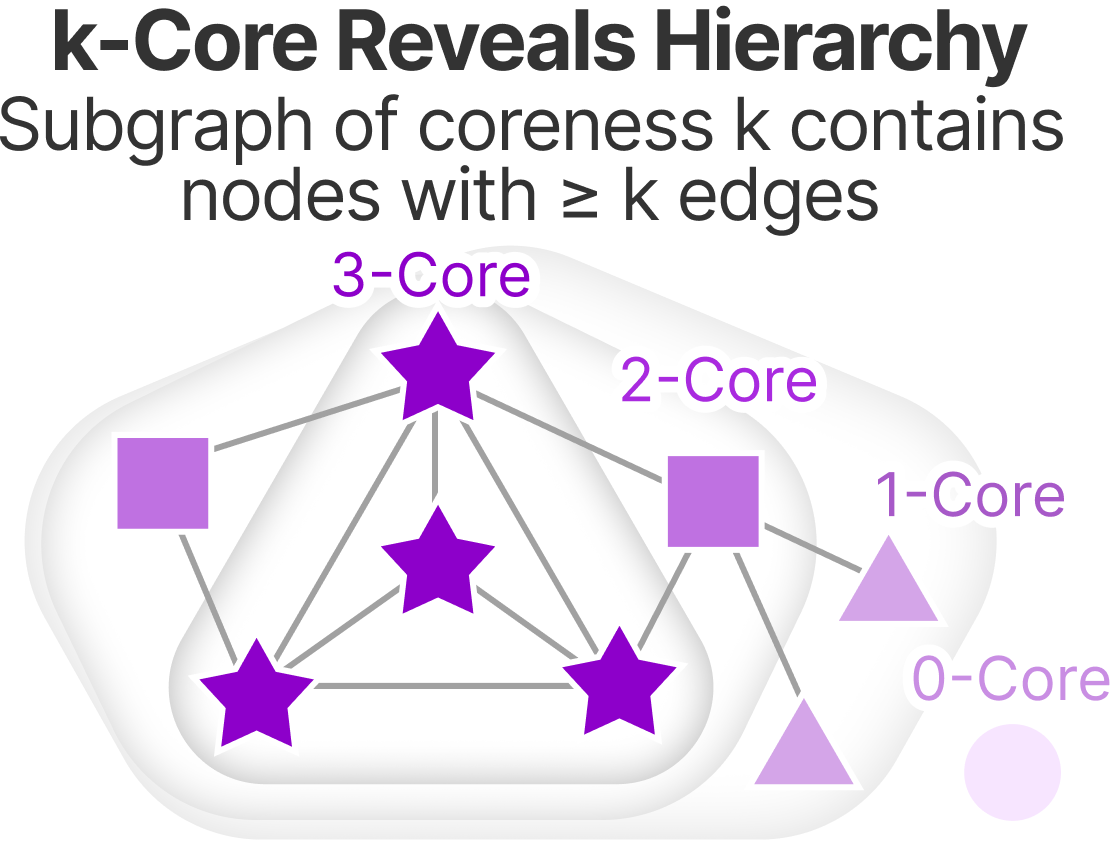}
\end{wrapfigure}

k-core decomposition~\cite{batagelj2003an} on the kNN graph provides a promising alternative. 
As shown in the right figure, iteratively removing nodes (and their edges) with the lowest degree in the remaining subgraph would assign each node a \textit{coreness} number (0, 1, 2, 3) which is the threshold at which it is removed. 
Higher core numbers indicate nodes embedded in progressively denser, more interconnected subgraphs. Lower numbers indicate more periphery.
A subtlety of $k$NN graphs is that every node has exactly $k$ outgoing edges, so any decomposition based on total degree (in + out) begins at a floor of $k$ and cannot distinguish truly peripheral nodes from moderately connected ones. We therefore decompose by in-degree alone (the number of other points that nominate a node as a neighbor) which varies naturally from 0 to well above $k$, producing a more discriminative hierarchy.

\subsubsection{Evaluation: k-Core}

\textbf{Structural characterization.}
On MNIST, coreness ranged from 0 to 8 (with $k{=}15$), with $65\%$ of points in the middle shells (4--6) and only $3\%$ peripheral.
Fashion MNIST showed greater structural heterogeneity: $18\%$ of points fell in the outermost shell.
Per-class analysis revealed that structurally simple, self-similar classes occupy the innermost core: on MNIST, the highest shell contained exclusively digit-1 images, the class with the least visual variation.
On Fashion MNIST, the innermost core was multi-class (Trouser~$29\%$, Pullover~$27\%$, Coat~$13\%$), indicating interleaved dense regions.
\Cref{fig:crownjewel}B shows examples of k-core decomposition uncovering distinct sub-categories in the dense ``bag'' points (e.g., messenger bags, waist packs, heavy textures) by filtering for coreness = 6.

\textbf{Comparison with HDBSCAN.}
We compared coreness against HDBSCAN membership probabilities using the recommended UMAP+HDBSCAN pipeline by McInnes.\footnote{\url{https://umap-learn.readthedocs.io/en/latest/clustering.html}. UMAP  (\texttt{n\_neighbors=30}, \texttt{min\_dist=0}) $\rightarrow$ HDBSCAN (\texttt{min\_cluster\_size=500}, \texttt{min\_samples=10}) on 2D projection.} 
HDBSCAN assigned nearly all points to clusters (${<}1\%$ noise), but its membership probabilities became nearly uniform across all coreness shells (mean~${\sim}0.93$, median~$1.0$ everywhere), yielding Spearman $\rho = 0.04$ on MNIST and $\rho = {-}0.01$ on Fashion MNIST.
This is expected: HDBSCAN answers \textit{which group?}, while k-core answers \textit{where in the group?}. Membership probability measures confidence that a point belongs to a cluster, not how central it is within it.
k-core, by contrast, produces a graduated hierarchy that distinguishes core from periphery within each cluster.

\subsection{Clustering Coefficient for Local Cohesion}
\label{sec:cc}

\setlength{\intextsep}{0pt}   %
\setlength{\columnsep}{10pt} %
\begin{wrapfigure}{r}{0.5\linewidth} %
  \centering
  \includegraphics[width=\linewidth]{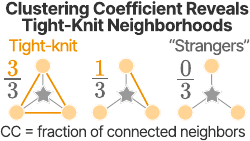}
\end{wrapfigure}

The \emph{clustering coefficient} (CC)~\cite{watts1998collective} acts as a microscopic lens, identifying tight-knit local neighborhoods.
Each node's CC ($[0, 1]$) is the fraction of ordered pairs of its nearest neighbors joined by a directed edge, e.g., 
CC is $1$ if every ordered pair of neighbors is directly connected; $0$ if none are.
A tight-knit region thus consists of points whose neighbors are themselves densely linked by nearest-neighbor nominations in high-dimensional space, yielding high CC.

\subsubsection{Evaluation: Clustering Coefficient}

On both datasets, CC was moderately correlated with k-core coreness (Spearman $\rho = 0.45$ on MNIST, $0.54$ on Fashion MNIST), PageRank ($0.44$, $0.51$), and local density ($0.38$, $0.38$), confirming CC captures related but non-redundant structure.
By contrast, CC was nearly uncorrelated with HDBSCAN membership probabilities ($\rho = 0.02$ on MNIST, $-0.05$ on Fashion MNIST), again reflecting that HDBSCAN measures cluster membership confidence, a fundamentally different question from local neighborhood cohesion.

\begin{figure}[tb]
\centering
\includegraphics[width=0.85\linewidth]{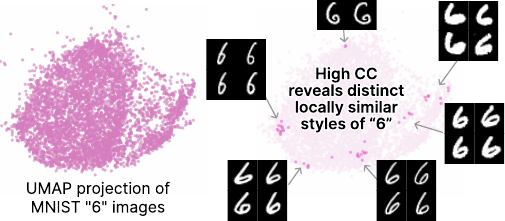}
\caption{Clustering coefficient (CC) isolates tight-knit handwriting styles of ``6'' (e.g., variations in tilt, loop size, stroke curvature).}
\label{fig:cc-mnist}
\vspace{-15pt}
\end{figure}

CC increased monotonically with coreness shell on both datasets: from $0.19$ in the outermost shell to $0.44$ in the innermost core on MNIST, and from $0.26$ to $0.75$ on Fashion MNIST.
This relationship shows that k-core and CC are complementary: k-core identifies macro-scale cohesive subgraphs, while CC pinpoints micro-scale cliques within those subgraphs.
Per-class analysis reinforced the structural narrative: digit~1 on MNIST had the highest mean CC ($0.376$), consistent with its role as the tightest, most self-similar class, while digit~8 had the lowest ($0.234$).
Quantitatively, the top 5\% of points by CC had mean neighborhood label purity of $0.98$ (MNIST) and $0.94$ (Fashion MNIST), vs.\ $0.90 \pm 0.004$ and $0.84 \pm 0.005$ for random same-size samples (100 runs), confirming high CC identifies semantically coherent micro-neighborhoods, not just topological cliques.
As illustrated in \Cref{fig:cc-mnist}, 
a UMAP projection of digit ``6'' appears as a monolithic mass, but filtering for high CC isolates distinct micro-neighborhoods of consistent handwriting styles (e.g., variations in tilt, loop size, stroke curvature).

\section{Implementation}
\label{sec:implementation}

All three techniques reuse UMAP's precomputed kNN arrays (indices, 
raw distances), avoiding neighbor recomputation.
PageRank additionally converts these distances into edge weights
via UMAP's membership-strength computation\footnote{Each
distance is normalized by per-point $\sigma_i / \rho_i$ parameters
computed by UMAP's \texttt{smooth\_knn\_dist} \cite{mcinnes2018umap-software}.}; k-core and clustering coefficient operate on the unweighted
directed graph.

\textbf{Computational cost.}
PageRank: $O(nk \cdot T)$ where $T$ is the number of power-iteration steps
(typically ${<}100$).
k-core: $O(nk)$.
Clustering coefficient: $O(nk^2)$.
All are fast relative to UMAP's own $O(nk)$ neighbor search and
$O(n \cdot \mathit{epochs})$ layout optimization.
On MNIST ($n{=}60{,}000$, $k{=}15$), each technique's wall-clock run time is under $1$\,s:
PageRank~$0.8$\,s, k-core~$0.6$\,s, clustering coefficient~$0.7$\,s.
(Re-deriving UMAP's membership strengths adds $0.9$\,s, which would vanish if UMAP exposed them.) All experiments were conducted on a MacBook Pro M4 Max with 64\,GB RAM.

\section{Conclusion and Discussion}
\label{sec:discussion}

We have shown that UMAP's internal kNN graph, typically set aside after layout optimization, is a rich analytical resource: PageRank, k-core decomposition, and clustering coefficient applied to it yield sensemaking complementary to the 2D scatter plot.
These methods inherit UMAP's $k$; PageRank rankings are highly stable across a $20{\times}$ range ($\rho \geq 0.84$), with moderate stability for coreness ($\rho \geq 0.89$) and CC ($\rho \geq 0.84$) across adjacent values (e.g., $k{=}15$ vs $30$).
PageRank has been integrated into the open-source Embedding Atlas~\cite{ren2025embedding} visualization, with k-core and clustering coefficient to follow.
The proposed graph-based perspective extends naturally to other DR methods that construct kNN graphs (e.g., TriMap~\cite{amid2019trimap}, PaCMAP~\cite{wang2021understanding}),
and graph algorithms beyond per-point scores (e.g., community detection~\cite{traag2019louvain,blondel2008fast}) are promising future directions.
Our evaluation focuses on two standard benchmarks; future work may extend to additional domains (e.g., single-cell genomics, audio).
We hope this work encourages the community to look beyond the scatter plot and into the graph that made it.

\acknowledgments{We thank our colleagues, especially Yannick Assogba and Ruisi Su, for their feedback.}

\bibliographystyle{abbrv-doi}

\balance
\bibliography{references}
\end{document}